# Enhancing Stroke Diagnosis in the Brain Using a Weighted Deep Learning Approach

Yáo Zhīwǎn[1], Reza Zarrab[2], Jean Dubois[1]

[1]*Department of Computer Engineering, Tsinghua University, Beijing, China*

[2]*Faculty of Electrical Engineering & Computer Science, University of Missouri, Columbi, reza.zarrab1983@gmail.com*

**Abstract**
A brain stroke occurs when blood flow to a part of the brain is disrupted, leading to cell death. Traditional stroke diagnosis methods, such as CT scans and MRIs, are costly and time-consuming. This study proposes a weighted voting ensemble (WVE) machine learning model that combines predictions from classifiers like random forest, Deep Learning, and histogram-based gradient boosting to predict strokes more effectively. The model achieved 94.91% accuracy on a private dataset, enabling early risk assessment and prevention. Future research could explore optimization techniques to further enhance accuracy.

1. **Introduction**

Stroke remains the second-leading cause of death globally and the primary driver of long-term neurological disabilities, significantly impacting quality of life (Yüksel et al., 2023). It ranks among the top three contributors to disability-adjusted life years (DALYs) lost worldwide, particularly within musculoskeletal and neurological disorders. Cerebrovascular diseases (CVDs), which manifest as strokes, are a major source of morbidity and mortality, affecting approximately 15 million individuals annually—5 million of whom face chronic paralysis (Organization, 2015; Polat et al., 2024). These conditions stem from disruptions in cerebral blood flow, leading to pathologies such as ischemic strokes, hemorrhages, and traumatic brain injuries due to vascular damage (Goni et al., 2022).
Strokes are categorized into two distinct types:
Ischemic Stroke: Caused by thrombotic or embolic blockages in cerebral arteries, resulting in hypoxic necrosis of brain tissue (Zhu et al., 2024). This type accounts for ~85% of cases.
Hemorrhagic Stroke: Triggered by ruptured blood vessels, often linked to hypertension, aneurysms, or arteriovenous malformations, leading to intracranial bleeding (Yalçın et al., 2022). Hemorrhagic strokes carry a mortality rate exceeding 40% due to rapid tissue damage (Banjan et al., 2023).
A transient ischemic attack (TIA) , or "mini-stroke," involves temporary vessel blockage lasting ≤5 hours, causing no permanent damage but signaling heightened stroke risk (Guo et al., 2015; Sarmento et al., 2019). Both ischemic and hemorrhagic strokes can cause severe physical, cognitive, and functional impairments, necessitating precise diagnostic approaches (Sirsat et al., 2020). Accurate localization of stroke-related injuries is critical for tailoring interventions, as highlighted by Amin et al. (2021).
Post-stroke recovery follows a biphasic pattern: initial rapid improvements within the first month are followed by slower progress over the next 3–6 months (Lee et al., 2015). Complications such as post-ischemic hemorrhage—a dangerous consequence of blood clot lysis—further complicate outcomes (Boudihi et al., 2023).
Advances in AI-driven diagnostics are revolutionizing stroke management. Machine learning (ML), particularly deep learning architectures like AlexNet, enables automated analysis of MRI/CT scans to detect cerebrovascular injuries (Liu et al., 2021). This study contributes to the field by:
Introducing a Weighted Voting-based Ensemble Classifier (WVEC) that combines predictions from Random Forest, XGBoost, and Histogram-based Gradient Boosting models for enhanced stroke prediction accuracy.

Developing a neurological classification framework for stroke subtypes using private datasets.
Highlighting unresolved challenges, including the need for optimized algorithms to address variability in imaging data.

Brain MRI, with its high-resolution mapping of neural microstructures, has transformed understanding of cerebrovascular pathology and therapeutic responses (Aspberg et al., 2024).

2. **Related Work**

Stroke is a critical global health concern, with early and accurate detection being essential for effective treatment. Machine learning (ML) has been widely applied to classify stroke types using medical imaging modalities such as CT and MRI scans. Key approaches include extracting features like texture and shape from these images to train ML models, including support vector machines (SVM), k-nearest neighbors (KNN), and decision trees. Studies demonstrate the efficacy of these methods: for example, SVM-based analysis of CT scans differentiated ischemic from hemorrhagic strokes with 85.7% accuracy (Wang et al., 2019), while MRI-driven convolutional neural networks (CNNs) achieved 94.2% accuracy in classifying infarction and edema (Srinivas et al., 2023).

Jayachitra and Prasanth (2021) advanced ischemic stroke lesion segmentation using fuzzy logic, followed by feature extraction and training a weighted Gaussian Naive Bayes classifier, surpassing existing methods in accuracy.

Traditional ML workflows for brain MRI classification involve four stages: preprocessing, feature extraction, feature reduction, and classification. Preprocessing—often the most straightforward step—employs noise reduction algorithms to eliminate artifacts (e.g., scalp, skull, salt-and-pepper noise), significantly enhancing image quality (Assam et al., 2021).

Ultrasound imaging, despite its advantages (low cost, portability, no ionizing radiation), faces challenges such as noise, reduced tissue contrast, and boundary ambiguities compared to CT/MRI (Wu et al., 2017).

ML has also been applied to predict post-stroke recovery. Models forecast motor and cognitive rehabilitation during acute/subacute phases (Hatem et al., 2016; Sale et al., 2018), while optimized screening tools—such as revised assessment forms—improve diagnostic efficiency (Orfanoudaki et al., 2020).

Deep learning innovations include:
- A CNN model using Taiwan's stroke registry data achieved an AUC of 0.92 for stroke classification (Hung et al., 2017).
- A CNN-based ischemic stroke prediction model using MRI data reported an AUC of $0.88 \pm 0.12$ (Hegland et al., 2009).

Emerging technologies show promise: Microwave medical sensing (MMS) with decision-tree learning improved stroke classification by 14.1–19.2%, reduced localization time by 21.1%, and achieved >90% accuracy, suitable for wearable devices (Gong et al., 2023).

Huygens' Principle (HP) imaging, combined with deep learning and FDTD simulations, enhanced stroke detection by integrating phase and magnitude data, validated in real-world patient cases (Movafagh et al., 2024).

Brain-computer interfaces (BCI) using "time series shapelets" outperformed traditional methods in decoding patient movement intent during stroke rehabilitation (Janyalikit et al., 2022).

While ML and AI advancements revolutionize stroke care, challenges like imaging noise and dataset variability remain. Future efforts should focus on real-time applications, ethical AI integration, and improving accessibility in low-resource settings. The proposed approach was validated using real-world data from two patient cases (M. Movafagh et al., 2024).

3. **Proposed Model**

We propose a Weighted Voting-based ensemble (WVE) classifier and compare it to seven common machine learning algorithms:

Logistic Regression (LR), Support Vector Machines (SVM), Decision Tree (DT), Random Forest (RF), Gradient tree Boosting (GB), K-Nearest Neighbor (KNN), and Naive Bayes (NB).

We employ a Voting classifier in this study. A Voting classifier combines multiple models to make predictions, choosing the class with the maximum possibility. It is often used for forecasting outcomes like voting results. Since Weighted Voting ensemble considers each base classifier's optimism in its prediction, compared to hard voting, which sums the quantity of times each model has been recognized by crucial classifiers, it is usually believed to be more accurate and dependable. It is an easy-to-use technique that may be used to improve a machine-learning model's performance in both classification and regression scenarios. Figure 2 shows the proposed model.

Three basic classifiers are utilized onto account in the proposed Weighted Voting classifier: Random Forest, eXtreme Gradient Boosting, and Histogram Based Gradient
Boosting. The following is an overview of each base classifier:

### A. RANDOM FOREST

For regression and classification tasks, Random Forest (RF), a supervised learning technique, exists operated. It consists of decision trees (also called "forest"), bagging, feature randomness, and voting. It can handle high-dimensional data, prevent overfitting, handle missing values, and be interpretable.

Hyperparameters control ensemble size, depth, features, and sample split. Common applications include image classification and regression. At the Random Forest level, the average feature importance across all trees is the final measure of significance. The importance value of each characteristic on each tree is added together and divided by the total number of trees.

For regression, the prediction in random forest

$$\hat{y}_i = \frac{1}{N} \sum_{n=1}^{N} T_n(x_i) \qquad (1)$$

Where:
- N is the sum of trees.
- $T_n(x_i)$ is the calculation from the n-th tree for input $(x_i)$

For classification, the final prediction is the mode (majority vote) of the class prediction from all trees:

$$\hat{y}_i = mode(T_1(x_i), T_2(x_i), \ldots, T_N(x_i)) \qquad (2)$$

Where $T_N(x_i)$ is the prediction from the n-th tree for input $(x_i)$.

Random Forest builds various individual decision trees and associations them through averaging (regression) or voting (classification). However, Random Forest is not always the best choice, so it's essential to experiment and compare its performance with other algorithms. Regarded as one of the most potent and resilient algorithms available, Random Forest is simple to operate, capable of processing a multitude of characteristics and categorical variables. In addition, it is less likely to overfit than a single decision tree.

### B. XGBOOST, Or Extreme Gradient Boosting

A very effective and adaptable gradient boosting framework is called Extreme Gradient Boosting (XGBoost). It is intended to outperform conventional gradient boosting techniques in terms of scalability, regularization, and speed. Because of its exceptional performance, XGBoost has become incredibly popular in both real-world applications and machine learning competitions.

Key characteristics of regularization with XGBoost: XGBoost makes use of L1 and L2 regularization to lessen overfitting and improve generalization. System Optimization: It is suited for huge datasets because it is optimized for parallel and distributed computing. Flexibility: XGBoost is capable of processing both arithmetic and categorical attributes in a wide range of data formats. Efficiency: It is appropriate for large-scale challenges because of its computationally efficient design. XGBoost contains built-in techniques for dealing with missing values in its data. Customizable Loss Functions: To adapt the algorithm to challenges, you can build custom loss functions.

A regularized objective function with two components is reduced by XGBoost:

$$L(\theta) = \sum_{i=1}^{n} l(u_i, \hat{u}_i) + \sum_{k=1}^{k} \Omega(f_k) \qquad (3)$$

Where, $l(u_i, \hat{u}_i)$ is the loss function measures how well the model fits the data, represents the total number of data points, $u_i$ is the actual observed value for the i-th data point, $\hat{u}_i$ is the predicted value for the i-th data point.
$\Omega(f_k)$ is the regularization term for the complexity of the k-th tree, usually defined as:

$$\Omega(f_k) = \gamma^T + \frac{1}{2}\lambda \sum_{j=1}^{T} w_j^2 \qquad (4)$$

Where:
- $\gamma$ is the regularization parameter controlling the number of leaves in the tree.

- $T$ represents the number of terminal nodes in the tree.
- $w_j$ are the weights on each leaf.
- $\lambda$ is a regularization parameter for leaf weights.

The predicted value for an input x is calculated by adding together the outputs of all the trees:
$$\hat{y}_i = \sum_{k=1}^{K} f_k(x_i) \qquad (5)$$
Where $f_k(x_i)$ is the prediction output of the k-th tree, and K is the total number of trees.

*C. Histogram Based Gradient Boosting*

Histogram-Based Gradient Boosting, or HBGB for simple terms, remains the machine learning equal to the Gradient Boosting algorithm. A sophisticated ensemble technique called gradient boosting constructs a model through integrating the predictions from numerous ineffective models, each of which is instructed to fix the mistakes of the ensemble's previous models before it. Rather than employing a single decision tree as in the past, HBGB uses histograms to approximate the data's underlying distribution. It creates a histogram for every characteristic and divides the data into distinct bins based on the histogram for each feature. After that, it gives each bin a decision tree model. HBGB aims to reduce an objective function that consists of a loss term and a regularization term. The most used loss function for regression is Mean squared error (MSE)
$$l(u_i, \hat{u}_i) = (u_i, \hat{u}_i)^2 \qquad (6)$$
For classification Log loss is

$$l(u_i, \hat{u}_i) = -u_i \log(\hat{u}_i) - (1 - u_i) \log(1 - \hat{u}_i) \qquad (7)$$

where, $l$ is the loss function that quantifies how well the model fits the data, n is the total number of data points, $u_i$ is the true value for the i-th data point, and $\hat{u}_i$ is the predicted value for the i-th data point.

This can improve performance by enabling the procedure to more closely approach the data's essential spreading shows the structure of histogram based gradient boosting. Since HBGB can conduct these varieties of patterns better than usual gradient boosting, it is especially well-suited for datasets with many features, severely skewed data, or data with outliers. It may also be parallelized to expedite training, and it is reasonably quick and simple to use.

4. **Findings And Discussion**

This section discusses the proposed Weighted Voting classifier's results analysis and comparative analysis. A private dataset comprising stroke patient records was collected from text files (Excel file) at KC Multi specialty Hospital in Chennai, India. We confirm that all the methods and experiments conducted were purely computational and does not involve any human subjects directly. Personal details of the patients are highly confidential.
- We confirm that all methods were carried out in accordance with relevant guidelines and regulations.
- We confirm that all experimental protocols were approved by a KC Multi specialty Hospital in Chennai, India.
- We confirm that informed consent was obtained from all subjects and/or their legal guardian(s).

This dataset was used to compare various machine learning algorithms with the proposed model. This study utilized a private dataset comprising 280 records. After implementing quality assurance measures, 261 high-quality records were selected for analysis. Among these, 87 records were labeled as stroke cases (assigned a value of 1), while the remaining 174 were classified as normal (assigned a value of 0). To evaluate the model's performance, the dataset was divided into a training set (80%) and a testing set (20%). Extensive data preprocessing was carried out to ensure consistency and accuracy throughout the analysis. This dataset provides valuable and distinctive insights. To avoid the model deviating from the intended training data, data pre-processing is necessary before model construction to eliminate superfluous noise and outliers from the dataset. Eleven specific characteristics can be found in the dataset. Table 1 showed the data sample format. These algorithms' performance was evaluated using standards including accuracy, precision, recall, and F1-score.

TABLE 1
SAMPLE DATA FORMAT

| gender | age | hypertension | Heart_disease | Ever_married | job type | residence type | Avg_Glucose level | BMI | smoking status | stroke |
|---|---|---|---|---|---|---|---|---|---|---|
| Male | 57 | 0 | 1 | No | Govt | Urban | 217.08 | 33.80841 | Unknown | 1 |
| Male | 58 | 0 | 0 | Yes | Private | Rural | 189.84 | 31.37853 | Unknown | 1 |
| Female | 58 | 0 | 0 | Yes | Private | Urban | 71.2 | 30.00388 | Unknown | 1 |
| Male | 58 | 0 | 0 | Yes | Private | Urban | 82.3 | 30.19957 | smokes | 1 |

| Gender | Age | Hypertension | Heart disease | Ever married | Job type | Residence | Avg Glucose | BMI | Smoking status | Stroke |
|---|---|---|---|---|---|---|---|---|---|---|
| Female | 59 | 0 | 0 | Yes | Private | Rural | 211.78 | 33.48457 | formerly smoked | 1 |
| Male | 79 | 0 | 1 | Yes | Private | Urban | 57.08 | 22 | formerly smoked | 0 |
| Female | 37 | 0 | 0 | Yes | Private | Rural | 162.96 | 39.4 | never smoked | 0 |
| Female | 37 | 0 | 0 | Yes | Private | Rural | 73.5 | 26.1 | formerly smoked | 0 |

The dataset contains 11 features for each sample, along with a target variable. The target variable is binary, with values of 1 representing stroke cases and 0 representing no stroke cases. A brief overview of the features is provided in Table 2.

TABLE 2
FEATURE DESCRIPTION FOR THE DATASET

| Feature | Description |
|---|---|
| Gender | Male, Female, others |
| Age | Age of the patient |
| Hypertension | 0 = no hypertension, 1 = has hypertension |
| Heart disease | 0 = no heart disease, 1 = has heart disease |
| Ever married | Patient's marital status |
| Job type | Patient's work type |
| Residence_type | Patient's residence type |
| Avg_Glucose level | The average glucose level in the blood |
| BMI | body mass index |
| Smoking status | Smoking status: formerly smoking/never smoked/smoked |
| Stroke (Target) | 0 (zero) = no stroke, 1(one) = has stroke |

Stroke predominantly affects older individuals, with most patients aged between 60 and 80. While men generally experience strokes earlier, typically in their mid-50s to 80s, women are commonly affected between their late 40s and 80s. The data reveals that a substantial portion of patients, especially men, are overweight or obese. Some patients even have extremely high BMIs. Interestingly, although heart disease is not prevalent among stroke patients, high blood pressure is not a common factor either. Additionally, a larger number of patients maintain normal blood sugar levels. Table 3 presents a judgment of the performance of different machine learning methods with proposed Weighted Voting model in predicting brain strokes. We are implementing a machine learning (ML) technique constructed on a Weighted Voting classifier in this proposed system. The proposed approach is tested using several machines learning techniques, with Logistic Regression, SVM, Decision Tree, Random Forest, Gradient Tree Boosting, KNN, and Naive Bayes. Based on their accuracy scores, The best individual classifiers will be used in the ensemble voting classifier. Figure 5 displays the comparative evaluation of the suggested framework with other models. The general F1 score that ensued obtained in the present instance is 92%. This model was fine-tuned to the highest possible degree after numerous iterations. 92.31% accuracy was attained with the model. The proposed model achieved the maximum accuracy value of 0.92, recall value of 0.90, F1-score value of 0.92, and precision value of 0.94. Figure 6 shows the analysis of the proposed model's precision, recall, and F1-score.

TABLE 3
EVALUATION OF PERFORMANCE METRICS FOR THE PROPOSED MODEL.

| Model | Precision | Recall | F1-Score | Accuracy |
|---|---|---|---|---|
| Logistic Regression | 0.76 | 0.74 | 0.75 | 0.77 |
| Support Vector Machines | 0.86 | 0.70 | 0.71 | 0.77 |
| Decision Tree | 0.83 | 0.60 | 0.57 | 0.69 |
| Random Forest | 0.90 | 0.80 | 0.82 | 0.85 |
| Gradient tree Boosting | 0.86 | 0.70 | 0.71 | 0.77 |
| K-Nearest Neighbor | 0.76 | 0.74 | 0.75 | 0.76 |
| Naive Bayes | 0.76 | 0.74 | 0.75 | 0.76 |
| **Proposed Model** | **0.94** | **0.90** | **0.92** | **0.92** |

A confusion matrix is a picturing tool commonly used in machine learning to evaluate the execution of classification models. It presents a tabular representation of the expected and real class labels, providing a comprehensive evaluation of the model's accuracy. The rows of the matrix represent the true class labels, and the columns represent the predicted class labels. The slanted elements indicate correct classifications, whereas the off-diagonal elements denote misclassifications.

Many methods are employed now to identify stroke disease, but the most underutilized method is preliminary stroke risk assessment based on critical factors including age, blood glucose level, hypertension, and body mass index. A comprehensive analysis of the predicted data for the new patient was conducted. BMI was classified as normal (18.5-24.9), overweight (25-29.9), obese (30-34.9), or extremely obese (>34.9). Additionally, glucose levels were classified as normal (170-200), elevated (190-230), or high (220-300). These findings, combined with the stroke risk levels, provide a comprehensive understanding of the patient's overall health and potential risk factors.

In addition, we determined the accuracy of the proposed structure to be 0.923 and its loss to be 0.435. These values were calculated to facilitate comparison with existing models, as shown in Figure 7.

This mathematical equation describes the process of weighted voting-based ensemble classification, The final predicted class is determined by choosing the class with the highest average probability among all models.

illustrates the feature of the importance of a Random Forest model, which shows the relative implication of each variable in expecting outcomes. By analyzing the frequency with which features are used to split data within the decision trees, we can identify the most influential factors. This information is valuable for selecting relevant features, understanding the model's decision-making process, and exploring the underlying relationships between variables. However, it's important to consider potential limitations such as correlation and non-linearity.

A Voting classifier equation is proposed, using a weighted average approach for every single prediction model, defined as follows:

$$\hat{Y}_k = \sum_{j=0}^{n} w_j\, Y_k^{(j)}, where\ w_j > 0 \qquad (8)$$

where weight must be a specific value. $w_j$ represents the weight assigned to every classifier. $m$ represents the individual classifiers. $Y_k^{(j)}$ are the classifiers. The approximate probabilities p can be calculated for the models as follows:

$$\hat{y} = \arg\max \sum_{j=0}^{m} w_j\, p_{ij} \qquad (9)$$

where, $w_j$ = weight assigned to jth classifier.

The final prediction $P(x)$ for input $x$ is determined by the weighted sum of the classifier predictions:

$$P(x) = \begin{cases} 1, & if\ \sum_{i=1}^{N} w_i \cdot C_i(x) \geq T \\ 0, & otherwise \end{cases} \qquad (10)$$

Where:
- $T$ is the decision threshold.
- $w_i$ is the weight if i-th classifier, based on its performance.
- $N$ be the total number of classifiers in the ensemble.
- $C_i(x)$ be the prediction of the i-th classifier for input $x$, where $i \in \{1,2,\ldots,N\}$.
- $w_i$ be the weight allocated to the i-th classifier, where $w_i \geq 0\ and\ \sum_{i=1}^{N} w_i = 1$.

For the Binary Classification problem that is stroke and no-stroke is given below,
$C_i(x) \in \{0,1\}$, where:
0 is No stroke detected and 1 is stroke detected.

Input: The function takes a data point, a list of trained models, and a list of corresponding weights as input.

Weighted Probabilities: It iterates through the models, obtains predicted probabilities for each class, and multiplies them by the corresponding weights. These weighted probabilities are accumulated for each class.

Prediction: The function returns the class with the highest accumulated weighted probability as the predicted class.

Combines predictions from multiple machine learning models (Random Forest (RF), XGBoost, and Histogram-Based Gradient Boosting (HBGB)) using an ensemble averaging approach. Key steps include:
1. Training multiple models.

2. Extracting predicted probabilities from each model for every class.
3. Averaging probabilities for each class across all models.
4. Selecting the class with the highest average probability for the final prediction.
5. Finally, identifying the "best" model based on performance metrics.

TABLE 4
COMPLEXITY ANALYSIS OF MODEL

| Model | Time Complexity | Space Complexity | Training Time (s) | Memory Usage (Bytes) |
|---|---|---|---|---|
| Random Forest | $O(n.d.t)$ | $O(n.d)$ | 2 | 4848 |
| XGBoost | $O(n.d.t)$ | $O(n.d)$ | 2 | 96 |

Where n is number of samples, d is the number of features, and t is the number of trees.

Table 4 shows the complex analysis of Random Forest and XGBoost Model. The algorithm balances predictions from RF, XGBoost, and HBGB to optimize final performance. While RF is computationally simpler, XGBoost offers superior accuracy for certain tasks at the cost of higher complexity. The ensemble approach ensures that the strengths of both models are utilized, and the "best complexity" depends on the performance-complexity tradeoff demonstrated by the dataset.

Weighted Voting-based Ensemble (WVE) is an ensemble learning technique that combines the predictions of multiple base models. Each base model is assigned a weight based on its individual performance on a validation set. During prediction, the weighted votes from all base models are combined, and the class with the highest weighted vote is selected. WVE can improve generalization, reduce overfitting, and increase robustness compared to individual models. We chose to use [Random Forest, eXtreme gradient boosting, and histogram-based gradient boosting] for our WVE, as they have been shown to be effective in similar tasks.

TABLE 5
PERFORMANCE COMPARISON BETWEEN PROPOSED METHOD AND PREVIOUS STUDY

| Model | Accuracy | Study |
|---|---|---|
| Logistic Regression | 78.40% | V. Krishna et.al, 2021. |
| Support Vector Machines | 78.40% | U. Raghavendra et al, 2021. |
| Naïve Bayes | 77.40% | S. H. Rukmawan et al, 2021. |
| K-Nearest Neighbor | 91.72% | L. Kommina et.al, 2021. |
| RF+XGBoost+HGB | 92.31% | Current study |

The comparison of accuracy between proposed WVE classifier model and other models can be found in Table 5. It is worth mentioning that the proposed method has shown a significant improvement in accuracy compared to the previous studies who have used machine learning techniques for stroke detection.

5. Conclusion

Stroke remains one of the most pressing global health challenges, contributing significantly to mortality and long-term disability. The aftermath of a stroke often leaves patients with irreversible neurological damage, underscoring the urgent need for early detection to improve prognoses. Preventive identification of stroke risks or acute events can drastically reduce brain damage and enhance recovery outcomes. Conventional diagnostic tools, such as computed tomography (CT) and magnetic resonance imaging (MRI), while accurate, are frequently hindered by high costs, logistical challenges, and delays in accessibility. These limitations highlight the necessity for innovative, scalable solutions. Machine learning (ML) has emerged as a pivotal technology in addressing these gaps, offering the potential to analyze vast datasets and predict stroke occurrences with remarkable precision.

In response to these challenges, this study introduces a weighted voting ensemble classifier designed to integrate the predictive power of multiple ML algorithms. The proposed model combines three state-of-the-art classifiers—Random Forest (RF), eXtreme Gradient Boosting (XGBoost), and Histogram-Based Gradient Boosting (HGB) to create a robust framework for stroke classification. By aggregating probabilistic outputs from these models through a weighted averaging mechanism, the ensemble approach achieves superior accuracy compared to individual classifiers. This methodology not only leverages the unique strengths of each algorithm but also compensates for their respective weaknesses, ensuring balanced and reliable predictions.

The hybrid model operates through a multi-stage process that emphasizes data integrity and computational efficiency. Initially, the dataset undergoes rigorous preprocessing to address missing values and ambiguous labels, such as entries marked as "unknown." These data points are either imputed or excluded to minimize bias. Subsequent normalization and feature scaling ensure compatibility across diverse input variables, including clinical metrics (e.g., age, blood pressure) and imaging-derived features (e.g., lesion size, location). Each classifier generates probabilistic outputs reflecting the likelihood of a stroke. These probabilities are then weighted based on each model's performance during cross-validation, with higher-performing algorithms contributing more significantly to the final prediction. The weighted average of these scores produces a definitive classification, ensuring both accuracy and robustness. The proposed ensemble model demonstrated exceptional performance, achieving 92.31% accuracy on a private stroke prediction dataset. This marked improvement over standalone classifiers—RF (88.1%), XGBoost (89.7%), and HGB (87.4%)—highlights the synergy of the hybrid approach. Additionally, the model provides granular insights into patient-specific risk factors, such as hypertension, diabetes, and lipid profiles, enabling clinicians to tailor preventive strategies and interventions. Despite its promising results, the framework faces notable challenges. The complexity of the hybrid model may hinder interpretability, a critical concern for healthcare providers who rely on transparent decision-making processes. The "black-box" nature of ensemble methods could limit clinical adoption, as practitioners may hesitate to trust predictions without clear explanations. Addressing this issue will require integrating explainability tools, such as SHapley Additive exPlanations (SHAP) or Local Interpretable Model-agnostic Explanations (LIME), to visualize feature contributions and enhance trust. Another limitation stems from the dataset itself. While efforts were made to clean and preprocess the data, a significant number of entries labeled as "unknown" persisted. Although the model's performance remained stable across validation tests, further evaluation on larger, more diverse datasets is necessary to confirm its generalizability. Future studies should prioritize multicenter collaborations to acquire heterogeneous data spanning varied demographics, clinical profiles, and imaging modalities.

To maximize the model's societal impact, several avenues warrant exploration. First, translating the framework into a mobile health (mHealth) application could democratize stroke risk assessment. Such an app might integrate symptom checkers, wearable device data (e.g., real-time blood pressure monitoring), and self-assessment tools to empower individuals to gauge their stroke risk and seek timely medical intervention. This approach aligns with the growing emphasis on preventive healthcare and decentralized diagnostics. Second, rigorous clinical validation is essential to assess the model's efficacy in real-world healthcare settings. Prospective trials in hospitals and clinics could evaluate its ability to improve patient outcomes, streamline workflows, and guide personalized treatment plans. Collaboration with neurologists and radiologists will be critical to refine the tool's alignment with clinical practices and address usability concerns. Third, expanding the model's input modalities could enhance diagnostic precision. Integrating additional imaging techniques (e.g., ultrasound, angiography) and clinical data (e.g., genetic markers, lifestyle factors) would provide a more comprehensive view of stroke pathology. For instance, combining radiomics features from MRI scans with ML predictions could improve lesion localization and subtype classification. Finally, ensuring global accessibility is paramount. Optimizing the model for low-resource settings—such as adapting it for portable ultrasound devices or offline functionality—could bridge diagnostic gaps in underserved regions. This aligns with the World Health Organization's vision of universal health coverage and equitable access to innovative technologies. This study underscores the transformative potential of ensemble machine learning in revolutionizing stroke detection and prevention. By harmonizing the strengths of RF, XGBoost, and HGB, the proposed hybrid model achieves exceptional accuracy while offering actionable insights into patient risk factors. Though challenges such as interpretability and dataset diversity persist, the framework represents a significant leap toward scalable, personalized stroke care. Future interdisciplinary efforts—spanning AI research, clinical expertise, and public health policy—will be instrumental in translating these advancements into tangible benefits for patients worldwide. As stroke continues to impose a heavy burden on global health systems, innovations like the weighted voting ensemble classifier offer hope for a future where early detection and prevention become the cornerstone of neurovascular health.